\documentclass[conference]{IEEEtran}
\usepackage{times}

\usepackage[numbers]{natbib}
\usepackage{multicol}
\usepackage[bookmarks=true]{hyperref}

\usepackage{graphicx}
\usepackage{amsmath, amssymb, amsthm}

\usepackage{algorithm, algorithmicx}
\usepackage[noend]{algpseudocode}

\usepackage{subcaption}
\usepackage{booktabs}

\newtheorem{property}{Property}[section]

\begin{document}

\title{Elaborating on Learned Demonstrations with Temporal Logic Specifications}


\author{\authorblockN{Craig Innes}
\authorblockA{School of Informatics\\
University of Edinburgh, UK\\
Email: craig.innes@ed.ac.uk}
\and
\authorblockN{Subramanian Ramamoorthy}
\authorblockA{School of Informatics \\
University of Edinburgh, UK\\
Email: s.ramamoorthy@ed.ac.uk}}


%

\newcommand{\ltl}{\textsc{ltl}}
\newcommand{\dmp}{\textsc{dmp}}
\newcommand{\lfd}{\textsc{lfd}}
\newcommand{\lc}{\mathcal{L}_{c}}
\newcommand{\xin}{\mathcal{I}}

\maketitle

\begin{abstract}

Most current methods for learning from demonstrations assume that those demonstrations alone are sufficient to learn the underlying task. This is often untrue, especially if extra safety specifications exist which were not present in the original demonstrations. In this paper, we allow an expert to elaborate on their original demonstration with additional specification information using linear temporal logic (\ltl{}). Our system converts \ltl{} specifications into a differentiable loss. This loss is then used to learn a dynamic movement primitive that satisfies the underlying specification, while remaining close to the original demonstration. Further, by leveraging adversarial training, the system learns to robustly satisfy the given \ltl{} specification on unseen inputs, not just those seen in training. We show that our method is expressive enough to work across a variety of common movement specification patterns such as obstacle avoidance, patrolling, keeping steady, and speed limitation. In addition, we show how to modify a base demonstration with complex specifications by incrementally composing multiple simpler specifications. We also implement our system on a PR-2 robot to show how a demonstrator can start with an initial (sub-optimal) demonstration, then interactively improve task success by including additional specifications enforced with a differentiable \ltl{} loss.
\end{abstract}

\IEEEpeerreviewmaketitle

\section{Introduction}
\label{sec:introduction}

Giving physical demonstrations is an established method of teaching a robot complex movements \cite{Argall2009ADemonstration}. However, such Learning from Demonstration (\lfd) techniques often assume the expert's demonstrations alone contain enough information to completely learn the underlying task. This is often untrue for at least two reasons:

First, the task may be difficult to demonstrate. This may be due to noisy sensors, control latency, fatigue, or lack of teleoperation skill. Simply copying the expert's movements may therefore be insufficient to perform the task adequately. For example, in a pouring task, the demonstrator might have been shaky and inadvertently spilled some of the liquid; on a peg insertion task, the demonstrator may overshoot the hole initially then re-adjust.

Second, and more importantly, many tasks have \emph{underlying} specifications which are simple to express verbally, but impossible to express through examples without an enormous amount of data \cite{Pinto2016SupersizingHours}. For example, if a reaching task has an easily formulated safety constraint such as ``don't move your hand above a certain speed when there is a fragile object nearby", it may take a large number of demonstrations to tease out this underlying behaviour.

For these reasons, we would ideally like to follow a model closer to human teaching --- one where we first give initial demonstrations, but can choose to elaborate on the precise detail of the task with additional instructions. In robotics, the language of Linear Temporal Logic (\ltl) \cite{Pnueli1977ThePrograms} provides us with an intuitive semantics with which to convey such instructions. \ltl{} is also expressive enough to capture most of the common specification patterns seen in typical robotic tasks \cite{Menghi2019SpecificationMissions}. Further, using \ltl{} opens up the potential for a user to add specifications directly with natural language, then later convert them into formal constraints using semantic parsing (we do not attempt this in this paper, but see e.g., \citet{Brunello2019SynthesisDirections} for details).

\begin{figure}
\centering
\includegraphics[width=\linewidth]{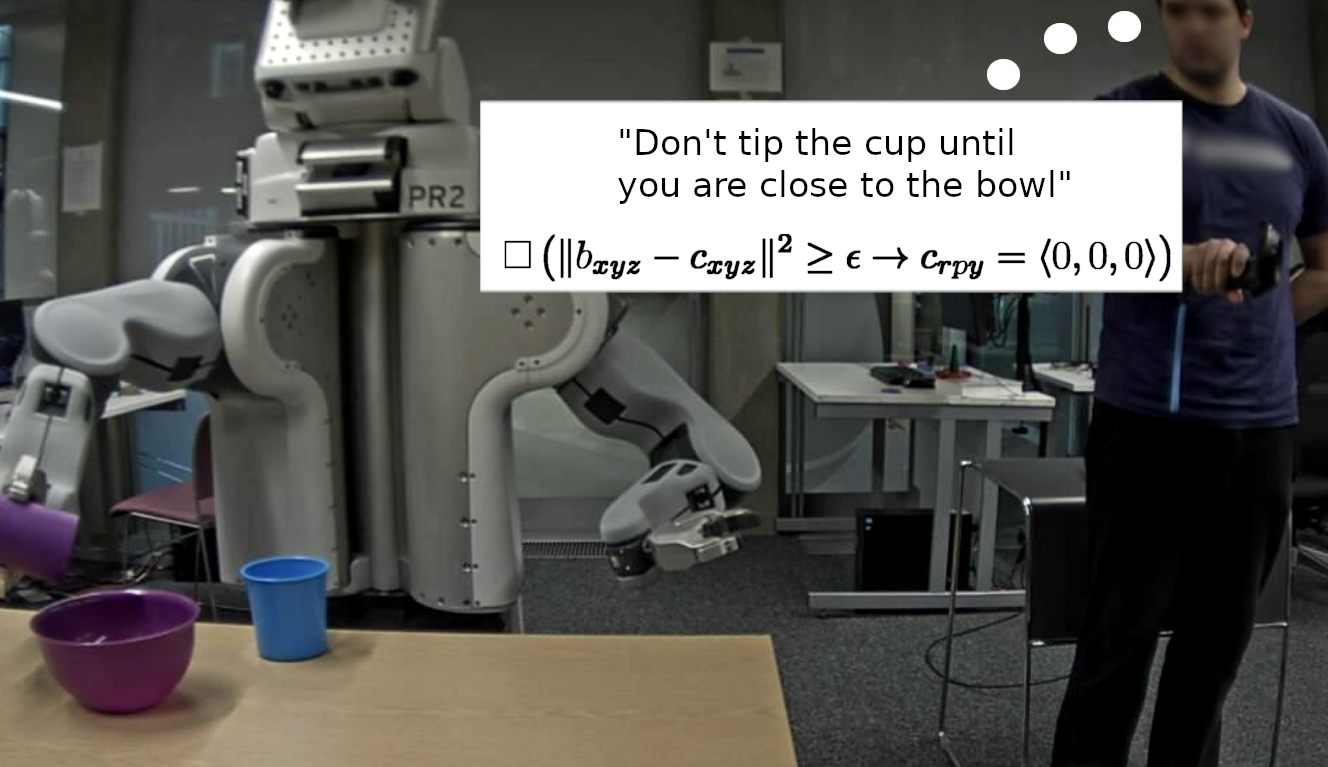}
\caption{A user pouring from a cup via teleoperation. The demonstration is later elaborated upon with the \ltl{} specification: ``Don't tip the cup until you are close to the bowl''.}
\label{fig:lfd-example}
\end{figure}

In this paper, we present a neural network architecture which learns to imitate expert demonstrations by learning the weights of a \emph{Dynamic Movement Primitive} (\dmp{}) \cite{Schaal2005LearningPrimitives} --- A generalizable representation for learning complex continuous movements. However, rather than merely optimizing the difference between the learned and target demonstrations, our system trades-off staying close to the original demonstration against satisfying additional \ltl{} specifications provided by the user. To achieve this, we provide a measure which transforms \ltl{} statements into a quantitative, differentiable metric. This means we can use this metric within the loss term of our optimization process. Further, by building on existing adversarial techniques for optimizing networks with simple propositional constraints (e.g., \citet{Fischer2019DL2:Logic}), our system can propose unseen counterexamples which are maximally likely to violate the temporal constraint given the current \dmp{}.

We incrementally build up task-learning by first learning a demonstration, then providing additional specifications later. In this sense, our work is close to the methodology of \emph{Interactive Task Learning} (\textsc{itl}). In such work, the goal is not just for the system to learn a given task, but to incrementally build an understanding of the task itself with help from the user.

In summary, our main contributions are:

\begin{enumerate}
    \item A loss function which transforms \ltl{} statements into a continuous, differentiable measure of satisfiability (Section \ref{sec:ltl}). Achieving a loss of zero guarantees that the constraint is satisfied for the given input.
    \item A model which takes in object locations and robotic poses, and outputs the weights of a \dmp{} (Section \ref{sec:dmps}). The model balances imitating the given demonstrations against satisfying the given \ltl{} constraints. By using adversarial learning, the model learns to satisfy constraints on not just the training data, but on unseen examples designed to violate the constraints (Section \ref{sec:adversarial-learning}).
    \item Experiments on 2-D curves across exemplary movement specifications such as obstacle avoidance, patrolling, staying within bounds, and staying within speed limits. These experiments combine demonstrations with task-relevant specifications to show two related uses of our system: First, we can elaborate on a single demonstration with \ltl{} to satisfy specifications not originally present (Section \ref{sec:one-shot-exp}). Second, we can take a model trained on a batch of demonstrations, and use our \ltl{} loss to satisfy a specification on an unseen test set (Section \ref{sec:generalized-exp}). 
    \item Application of our model to reaching and pouring tasks on a PR-2 robot. The inputs are the gripper pose and object locations (extracted using the PR-2's RGB-D camera). The output is a sequence of poses, which is synthesized into robotic motion using an inverse kinematics library. We show how a user can start with an initial demonstration captured via HTC-Vive VR controllers, then elaborate on that demonstration using \ltl{} statements to either provide hints about how to better satisfy the task, or to modify the original task specification (Section \ref{sec:pr2-exp}).
\end{enumerate}

\section{Combining lfd and ltl}
\label{sec:method}

In this section, we describe a method for teaching a robot to learn movements for \emph{finite horizon} tasks from a combination of expert demonstrations and specifications given in \ltl{}. We first briefly introduce learning to imitate demonstrations with \dmp{}s (\ref{sec:dmps}), as well as the basics of \ltl{} (\ref{sec:ltl}). Then, we outline our contributions, which are to formulate a differentiable loss function for \ltl{} statements, and then optimize this loss using adversarial training techniques (\ref{sec:adversarial-learning}).

\subsection{Learning from Demonstrations with DMPs}
\label{sec:dmps}

Lets assume we have data for several demonstrations $D = \{ D_0 \dots D_M \}$ in the form $D_i = \langle \xin_i, y_i \rangle$. Here, $\xin_i$ denotes the starting input for the demonstration, which may include relevant objects positions, the initial joint-angles/pose of the robot, or many other sensory inputs. The term $y_{i} = [y_{i, 0} \dots y_{i, T}]$ represents the \emph{trajectory} of the robot, where $y_{i, j}$ is the recorded pose at time $j$ of demonstration $i$.

The goal is to learn a function that, given input $\xin_i$, outputs trajectories that are close to the demonstrations. One common way to represent such trajectories is as a Dynamic Movement Primitive (\dmp{}). \dmp{}s allow us to compactly represent a trajectory as a series of differential equations:

\begin{align}
    \label{eqn:dmp-acceleration}
    \ddot{y} &= \alpha_y ( \beta_y ( y_{goal} - \dot{y}) - \dot{y}) + f(x) \\
    \label{eqn:dmp-canonical}
    \dot{x} &=  -\alpha_x x
\end{align}

The acceleration of the end effector, $\ddot{y}$, has two parts: The first is a point attractor, which linearly moves towards $y_{goal}$ at a rate determined by the distance ($y_{goal} - y$), a drag term ($-\dot{y}$), and scalars $\alpha_y$ and $\beta_y$. The second is the function $f(x)$, which expresses complex non-linear motion as a combination of radial basis functions $\psi_{0 \dots N}$, weighted by parameters $w_{0 \dots N}$:

\begin{equation}
    \label{eqn:dmp-rbfs}
    f(x) = \frac{\sum_{i=0}^{N} w_i \psi_i(x)}{\sum_{i=0}^N \psi_i(x)}x(y_{goal} - y_{start})
\end{equation}

Equations (\ref{eqn:dmp-acceleration}-\ref{eqn:dmp-rbfs}) introduce a structural bias which favours smooth progress towards a goal over arbitrary movement.

Rather than using time directly, these equations are governed by the \emph{canonical system} x, which starts at 1 and decays logarithmically towards 0.

With the above equations, we can now formally define the \lfd{} problem: Given a demonstration $D_i$, we wish to learn a parameterized function $g_{\theta}$ which takes $\xin_i$ as input and produces weights $w$ as outputs such that the following loss function is minimized:

\begin{equation}
    \label{eqn:loss-imitation}
    \mathcal{L}_{d}(\theta, \xin_i, y_i) = \frac{1}{T} \sum_{t = 0}^{T}\lVert \dmp_{\theta, \xin_i}(t) - y_{i,t} \rVert^2 
\end{equation}

Here, $\dmp_{\theta, \xin_i}(t)$ denotes the value at time $t$ of a \dmp{} defined by  $y_{start} = y_{i, 0}$, $y_{goal} = y_{i, T}$, and weights learned learned via a neural network function parmeterized by $\theta$:

\begin{equation}
    \label{eqn:dmp-weight-nn}   
    w = g_{\theta}(\xin_i) 
\end{equation}
    
In words, (\ref{eqn:loss-imitation}) gives the average distance between the learned and demonstrated trajectories. We call (\ref{eqn:loss-imitation}) the \emph{imitation loss}.

\begin{figure}
\centering
\includegraphics[width=\linewidth]{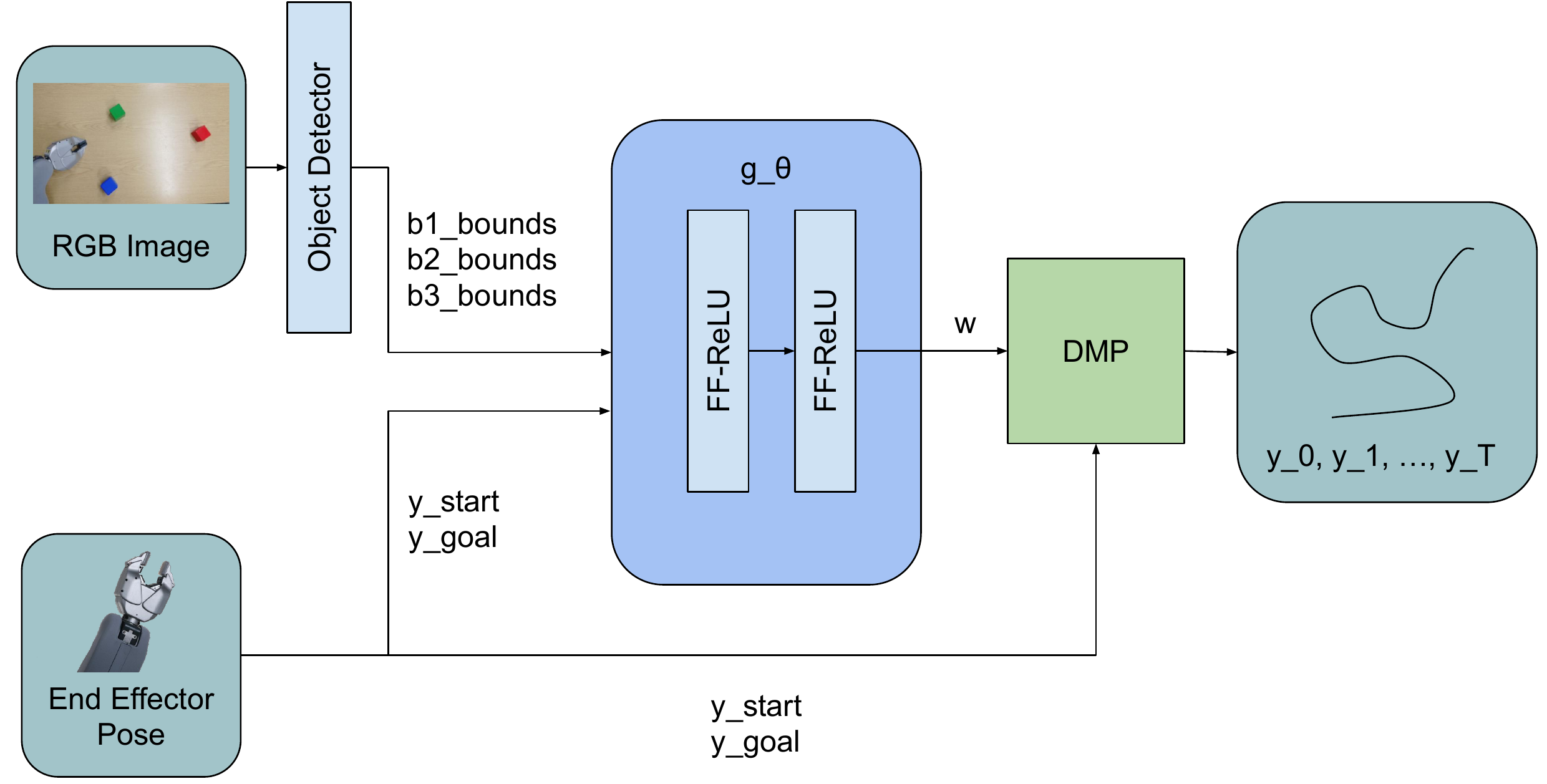}
\caption{Architecture overview. Inputs are the gripper pose and object descriptions (either given, or extracted from RGBD with an off-the-shelf object detector). We run these through two feed-forward layers with ReLU activations, and output weights $w$ for equation (\ref{eqn:dmp-rbfs}). To calculate the trajectory, we roll out equations (\ref{eqn:dmp-acceleration}-\ref{eqn:dmp-canonical}).}

\label{fig:architecture-overview}
\end{figure}

Figure \ref{fig:architecture-overview} shows an architecture overview. Like other works \cite{Pervez2017LearningNetworks, Pahic2018DeepPrimitives}, we represent $g_{\theta}$ as a feed-forward network, allowing us to optimize (\ref{eqn:loss-imitation}) using stochastic gradient-descent\footnote{For simple \dmp{}s,  we can approximate $w$ in closed-form \cite{Ijspeert2013DynamicalBehaviors}. However, such methods do not accommodate additional sensory inputs, nor the modifications we make to the loss function in the next section.}.

We now have a system which imitates demonstrated trajectories as closely as possible. The next sections outline how to extend (\ref{eqn:loss-imitation}) beyond mere imitation with additional specifications.

\subsection{A Loss Function from Temporal Logic}
\label{sec:ltl}

Linear Temporal Logic (\ltl{}) \cite{Pnueli1977ThePrograms} is a language which extends traditional propositional logic with modal operators. With these additional operators, we can specify robotic constraints that must hold \emph{through time}. The syntax for constructing an \ltl{} constraint $\varphi$ is given by (\ref{eqn:ltl-syntax}):

\begin{equation}
    \begin{aligned}
    \label{eqn:ltl-syntax}
    &\varphi := p \mid \neg \varphi \mid \varphi_1 \wedge \varphi_2 \mid \mathcal{N} \varphi \mid \square \varphi \mid \diamondsuit \varphi \mid \varphi_1 \mathcal{U} \varphi_2 \\
    &p := T_1 < T_2 \mid T_1 \leq T_2 \mid T_1 = T_2 \mid T_1 \geq T_2 \mid T_1 > T_2
    \end{aligned}
\end{equation}

Here, $\mathcal{N} \varphi$ means that $\varphi$ must hold at the \emph{next} time-step, $\square \varphi$ means that $\varphi$ must hold at \emph{every} time-step, $\diamondsuit \varphi$ means that $\varphi$ must \emph{eventually} hold at some future time step, and $\varphi_1 \mathcal{U} \varphi_2$ means that $\varphi_1$ must remain true \emph{until} $\varphi_2$ holds. Atomic comparisons $p$ may be made up of \emph{static} terms (which do not depend on time), or \emph{dynamic} terms (which do depend on time, such as the value of a \dmp{}). Terms are denoted by $T_i$.

Let's say we wanted to express the constraint: ``The robot must always stay away from obstacle $o_{bad}$, and must at some point visit $o_{good}$". We could express this in \ltl{} as:

\begin{equation}
   \label{eqn:ltl-example} 
   \left(\square \lVert \dmp_{\xin_i, \theta} - o_{bad} \rVert^2 \geq c \right ) \wedge \left( \diamondsuit \dmp_{\xin_i, \theta} = o_{good} \right) 
\end{equation}

We assume here that the positions of $o_{bad}$ and $o_{good}$ are static, while the current output of $\dmp_{\xin_i, \theta}$ is dynamic, as it depends on the current time-step $t$.

The \ltl{} definition given above is useful for discrete verification and high-level synthesis \cite{Kress-Gazit2011CorrectControl}. However, we cannot currently use it in our optimization process outlined in section \ref{sec:dmps} as it does not provide a continuous gradient to optimize with respect to. To use it for this purpose, we introduce a \emph{constraint loss} $\mathcal{L}_{c}$ which takes as input an \ltl{} statement $\varphi$ and time-step $t$, and outputs a real-valued, differentiable loss. This loss measures quantitatively how close $\varphi$ is to being satisfied.

Equation (\ref{eqn:ltl-diff-semantics-atomic}) gives the quantitative semantics for the atomic comparison terms. These build on the work by \citet{Fischer2019DL2:Logic} who define a differentiable quantitative semantics for propositional logic:

\begin{equation}
    \label{eqn:ltl-diff-semantics-atomic}
    \begin{aligned}
        &\mathcal{L}_c(t_1 \leq t_2, i) = max(t_1(i) - t_2(i), 0) \\
        &\mathcal{L}_c(t_1 \neq t_2, i) = \zeta  [ t_1(i) = t_2(i) ] \\
        &\mathcal{L}_c(t_1 < t_2, i) = \mathcal{L}_c(t_1 \leq t_2 \wedge t_1 \neq t_2, i) \\
        &\mathcal{L}_c(t_1 = t_2, i) = \mathcal{L}_c(t_1 \leq t_2 \wedge t_1 \geq t_2, i) \\
    \end{aligned}
\end{equation}

Here, $[a = b]$ is a function which evaluates to 1 when $a = b$ and 0 otherwise. The constant $\zeta$ is a small positive scalar.

Equation (\ref{eqn:ltl-diff-semantics-compose}) gives the quantitative semantics for the modal and composition operators. It builds on quantitative measures of robustness from signal temporal logic \cite{Fainekos2009RobustnessSignals}:

\begin{equation}
    \label{eqn:ltl-diff-semantics-compose}
    \begin{aligned}
        &\mathcal{L}_c(\varphi_1 \wedge \varphi_2, i) = max^{\gamma}\left(\mathcal{L}_c(\varphi_1, i), \mathcal{L}_c(\varphi_2, i)\right) \\
        &\mathcal{L}_c(\varphi_1 \vee \varphi_2, i) = min^{\gamma}(\mathcal{L}_c(\varphi_1, i), \mathcal{L}_c(\varphi_2, i)) \\
        &\mathcal{L}_c(\mathcal{N} \varphi, i) = \mathcal{L}_c(\varphi, i + 1) \\
        &\mathcal{L}_c(\square \varphi, i) = max^{\gamma}(\mathcal{L}_c(\varphi, i), \dots, \mathcal{L}_c(\varphi, T)) \\
        &\mathcal{L}_c(\diamondsuit \varphi, i) = min^{\gamma}(\mathcal{L}_c(\varphi, i), \dots, \mathcal{L}_c(\varphi, T)) \\
        &\mathcal{L}_c(\neg \mathcal{N} \varphi, i) = \mathcal{L}_c(\mathcal{N} (\neg \varphi), i) \\
        &\mathcal{L}_c(\neg \square \varphi, i) = \mathcal{L}_c(\diamondsuit (\neg \varphi), i) \\
        &\mathcal{L}_c(\neg \diamondsuit \varphi, i) = \mathcal{L}_c(\square (\neg \varphi), i) \\
        &\mathcal{L}_c(\varphi_1 \mathcal{U} \varphi_2, i) = max^{\gamma}(\\
            &\quad min^{\gamma}(\lc(\varphi_1, i), \lc(\varphi_2, i)),\\
            &\quad min^{\gamma}(\lc(\varphi_1, i + 1), \lc(\varphi_2, i), \lc(\varphi_2, i + 1)),\\
            &\quad \dots, \\
            &\quad min^{\gamma}(\lc(\varphi_1, T), \lc(\varphi_2, i), \lc(\varphi_2, i + 1), \dots, \lc(\varphi_2, T)))
    \end{aligned}
\end{equation}


Because maximization and minimization functions are non-differentiable, we use the soft approximations $max^{\gamma}$ and $min^{\gamma}$ (inspired by the work of \citet{Cuturi2017Soft-DTW:Time-series} on soft approximations for dynamic time warping). These are defined in equations (\ref{eqn:soft-max}) and (\ref{eqn:soft-min}) below:

\begin{gather}
    \label{eqn:soft-max}
    max^\gamma(x_1, \dots, x_n) = \gamma \ln{\sum_{i=0}^{n} e^{x_i / \gamma}}\\
    \label{eqn:soft-min}
    min^\gamma(x_1, \dots, x_n) = -\gamma \ln{\sum_{i=0}^{n} e^{-x_i / \gamma}}
\end{gather}

The above loss function has two useful properties. First, it is differentiable\footnote{To be precise, it is technically \emph{almost-everywhere} differentiable.}. Second, it has the following soundness property (by construction):

\begin{property}{Soundness of $\mathcal{L}_c$:}
Given $\gamma \rightarrow 0$, for any $\varphi$ and $i$, if $\mathcal{L}_c(\varphi, i) = 0$, then the constraint $\varphi$ is satisfied at time-step $i$ according to the original qualitative semantics of \ltl{}.
\end{property}

In words, this means that minimizing $\mathcal{L}_c$ is equivalent to learning to satisfy the constraint.

With the above properties, we can now augment our \lfd{} loss from (\ref{eqn:loss-imitation}) with \ltl{} constraints $\varphi$ to give equation (\ref{eqn:full-loss-train}):

\begin{equation}
    \label{eqn:full-loss-train}
    \mathcal{L}_{full}(\theta, D, \varphi) = 
    \frac{1}{M} \sum_{i = 0}^M \mathcal{L}_{d}(\theta, \xin_i, y_i) + \eta  \mathcal{L}_{c}(\varphi(\theta, \xin_i), 0)
\end{equation}

Here, we somewhat abuse notation and write $\varphi(\theta, \xin)$ to mean evaluating $\varphi$ with the values $\theta$, and $\xin$ substituted in as the values of the corresponding variables within the constraint. The term $\eta$ is a constant which decides how much to trade off between perfectly imitating the expert demonstrations, and ensuring that $\varphi$ is satisfied across all training inputs.

\subsection{Adversarial Training for Robust Constraint Satisfaction}
\label{sec:adversarial-learning}

By optimizing equation (\ref{eqn:full-loss-train}), we can (in the best case) learn parameters $\theta$ which output movements guaranteed to satisfy the user-specified constraints. However, such guarantees only hold \emph{for the training inputs} $\xin_{0\dots M}$. For scenarios in which we wish to augment the \emph{existing} behaviour of a \emph{single} demonstration, this is fine. However in scenarios where we are training a \emph{general} behaviour with a \emph{batch} of varying demonstrations, we cannot be sure of whether our model's ability to satisfy the given constraints will generalize to unseen examples.

This lack of generalization is a particular issue for \ltl{} constraints which represent important safety specifications. For such constraints, we are often concerned with not only the average case behaviour of our model, but how the model will behave in the \emph{worst possible situation} --- we want to ensure our model satisfies the constraint \emph{robustly}.

Let us clarify this notion of robustness with reference to the example from equation (\ref{eqn:ltl-example}): Ideally, our robot should avoid $o_{bad} = \xin_{i, j}$ and reach $o_{good} = \xin_{i, k}$ even if we slightly change all relevant inputs by some small amount $\epsilon$. We want to be \emph{robust} to perturbations, which we can write formally as equation (\ref{eqn:ltl-example-robust}):

\begin{equation}
    \begin{aligned}
    \label{eqn:ltl-example-robust}
    &\forall z_i \in B_{\epsilon}(\xin_i), \\
    &\quad \left(\square \lVert \dmp_{z_i, \theta} - z_{i, j} \rVert^2 \geq c \right ) \wedge \left( \diamondsuit \dmp_{z_i, \theta} = z_{i, k} \right) 
    \end{aligned}
\end{equation}

Here, $B_{\epsilon}(\xin) = \big\{ z \mid \lVert \xin - z \rVert^{\infty} < \epsilon \big\}$ is the set of all points which lie within $\epsilon$ of $\xin$ according to the $L_{\infty}$ norm.

It is impractical to exhaustively check (\ref{eqn:ltl-example-robust}) holds at every valid instantiation of $z_i$. However, we can still robustly enforce this constraint by leveraging the following equivalence noted in \citet{Fischer2019DL2:Logic}: Saying $\varphi$ holds everywhere is the same as saying there is no counter-example $z'$ where $\varphi$ \emph{does not hold}.

This insight suggests the following two-step adversarial training loop: First, have an adversary find a $z'$ which minimizes the \emph{negated} loss $\mathcal{L}_c(\neg \varphi, 0)$. Second, give this $z'$ as input to our network, and attempt to find the parameters $\theta$ which minimize the full loss $\mathcal{L}_{full}(\theta, D, \varphi)$.

Algorithm \ref{alg:full-system} outlines our training process. Lines (6-9) show the adversarial inner loop. First, we sample a $z$ from the region around the current training example. We then repeatedly run $z$ through our network $g_{\theta}$, and minimize the negated loss of our constraint $\varphi$ using projected gradient descent. In lines (10-11), we then take the $z$ chosen by our adversary, and use its gradient with respect to $\mathcal{L}_{full}$ to update our network parameters $\theta$.

\begin{algorithm}
        \caption{Adversarial training with demos and \ltl{}}
        \label{alg:full-system}
        \begin{algorithmic}[1]
        \Function{Train-DMP}{$D$, $\epsilon$, $\gamma$, $\eta$, $\varphi$, $\mathit{epochs}$, $\mathit{iterations}$}
            \State Initialize $\theta$
            \For{$1 \dots \mathit{epochs}$}
                \For{$\langle \xin_i, y_i \rangle \in D$}
                    \State $z \gets$ Sample from $B_{\epsilon}(\xin_i)$
                    \For{$1 \dots \mathit{iterations}$}
                        \State $y' \gets$ Rollout \dmp{} with weights $w = g_\theta(z)$
                        \State $z \gets$ Update using $\nabla_z \mathcal{L}_c(\neg \varphi(\theta, z), 0)$
                        \State $z \gets$ Project $z$ onto $B_{\epsilon}(\xin_i)$
                    \EndFor
                    \State $y \gets$ Rollout \dmp{} with weights $w = g_\theta(z)$
                    \State $\theta \gets$ Update with: 
                    \State \quad $\nabla_{\theta}\left(\mathcal{L}_{d}(\theta, \xin_i, y_i) + \eta \mathcal{L}_c(\varphi(\theta, z), 0)\right)$
                \EndFor
            \EndFor
        \Return $\theta$
        \EndFunction
        \end{algorithmic}
\end{algorithm}

In summary, we have a model which takes expert demonstrations combined with additional \ltl{} specifications as input, and outputs the weights of a \dmp{}. We can train this model to alter a single initial demonstration to obey additional constraints. Alternatively, we can leverage adversarial training and train our model to produce \dmp{}s which satisfy the given constraints not just on the training data, but on unseen examples in a robust region.

\section{Experiments}
\label{sec:experiments}

In this section, we show that the model presented in section \ref{sec:method} is practically useful via experiments across a range of tabletop movement tasks and across a variety of specifications. Our model is useful in two senses: First, we show that we can incrementally elaborate on the behaviour of a \emph{single} demonstration using additional \ltl{} specifications. Second, we show that given a \emph{batch} of demonstrations augmented with an \ltl{} specification, our adversarial learning approach outlined in algorithm \ref{alg:full-system} allows us to generalize the desired specification to unseen inputs.

We start by applying our model to 2-D planar motion scenarios. In subsequent sections we show example applications of our model on a PR-2 robot.

\subsection{One-Shot 2-D Tasks}
\label{sec:one-shot-exp}

\begin{figure*}
    \centering
    
    \includegraphics[width=0.3\textwidth]{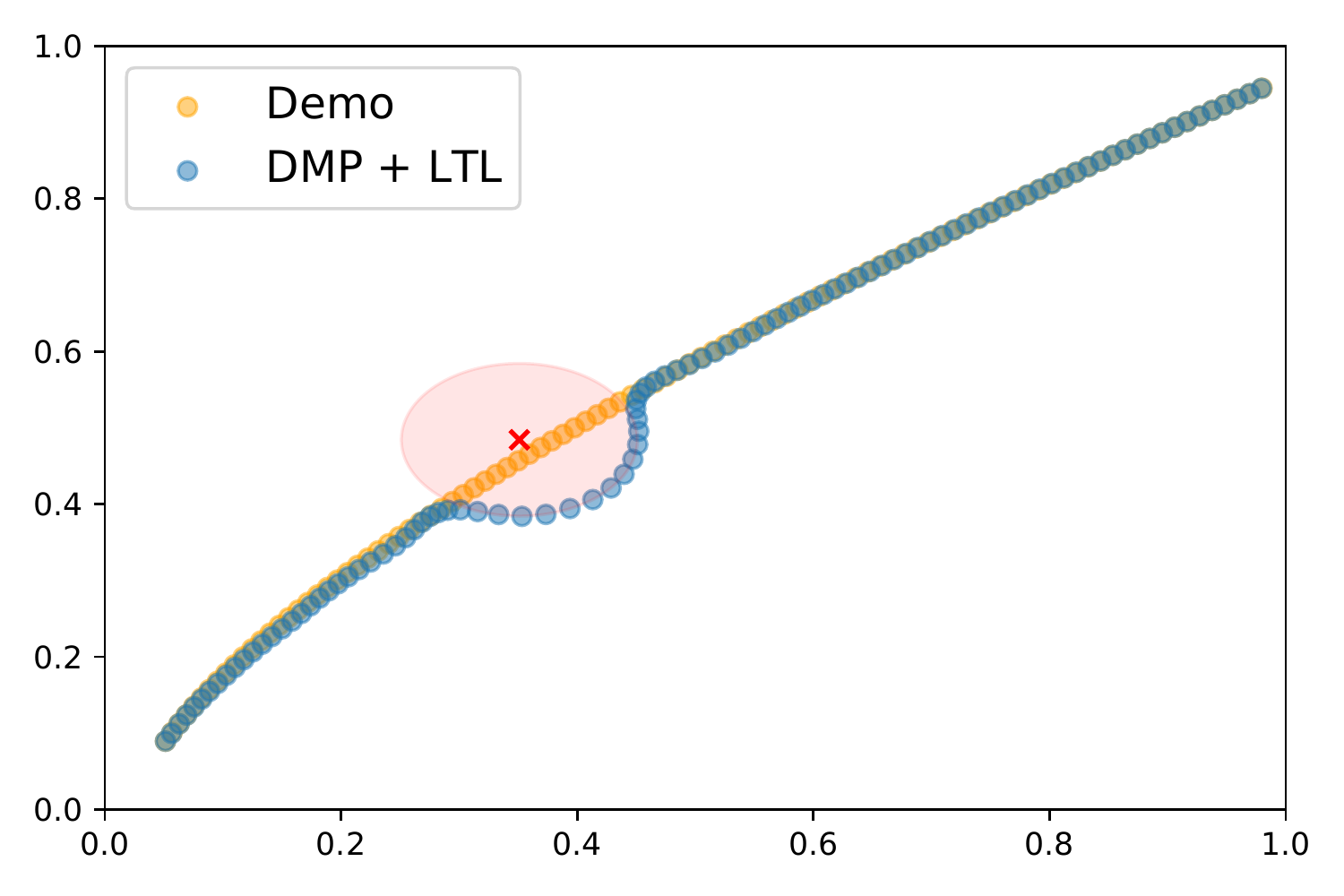}
    \includegraphics[width=0.3\textwidth]{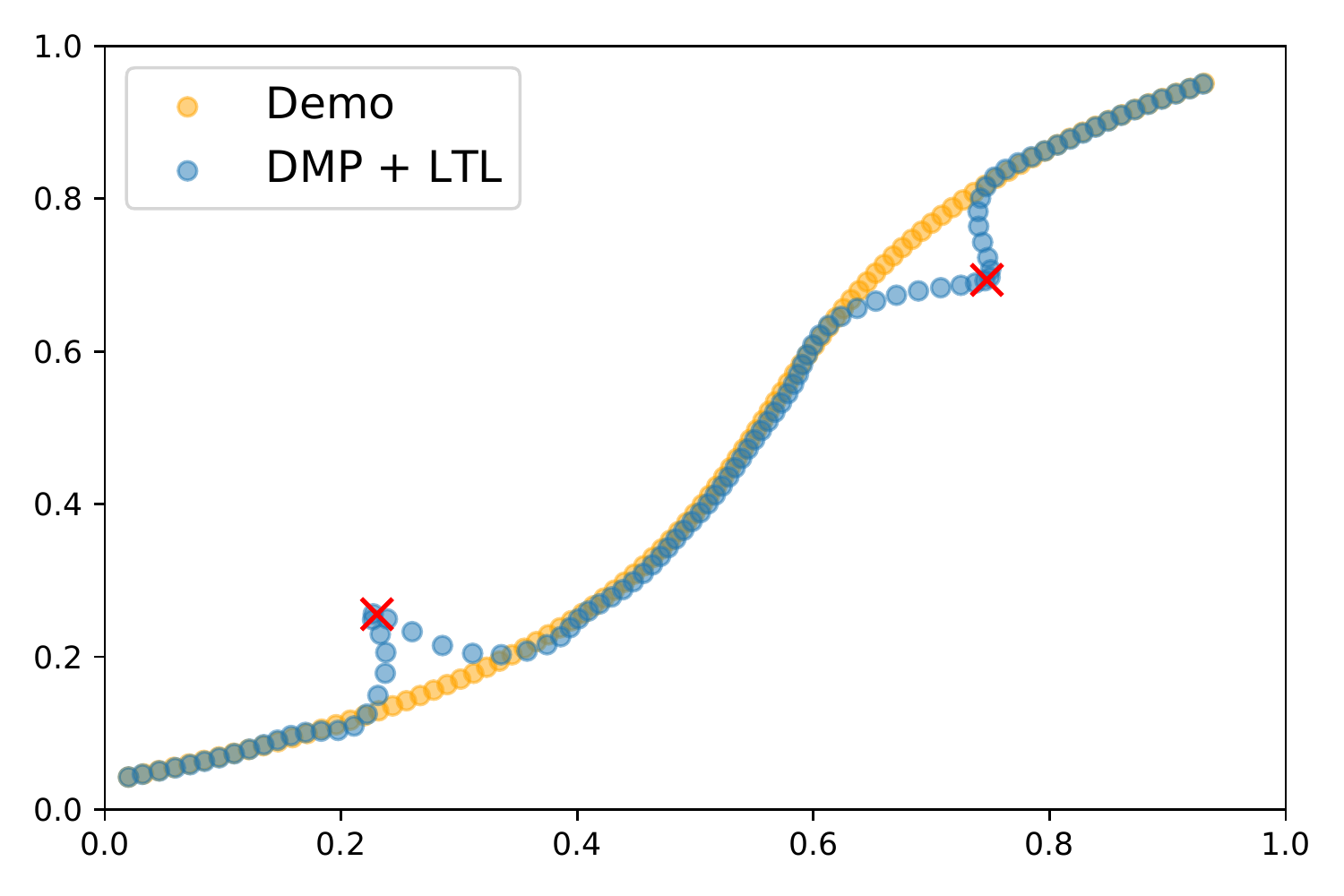}
    
    \includegraphics[width=0.3\textwidth]{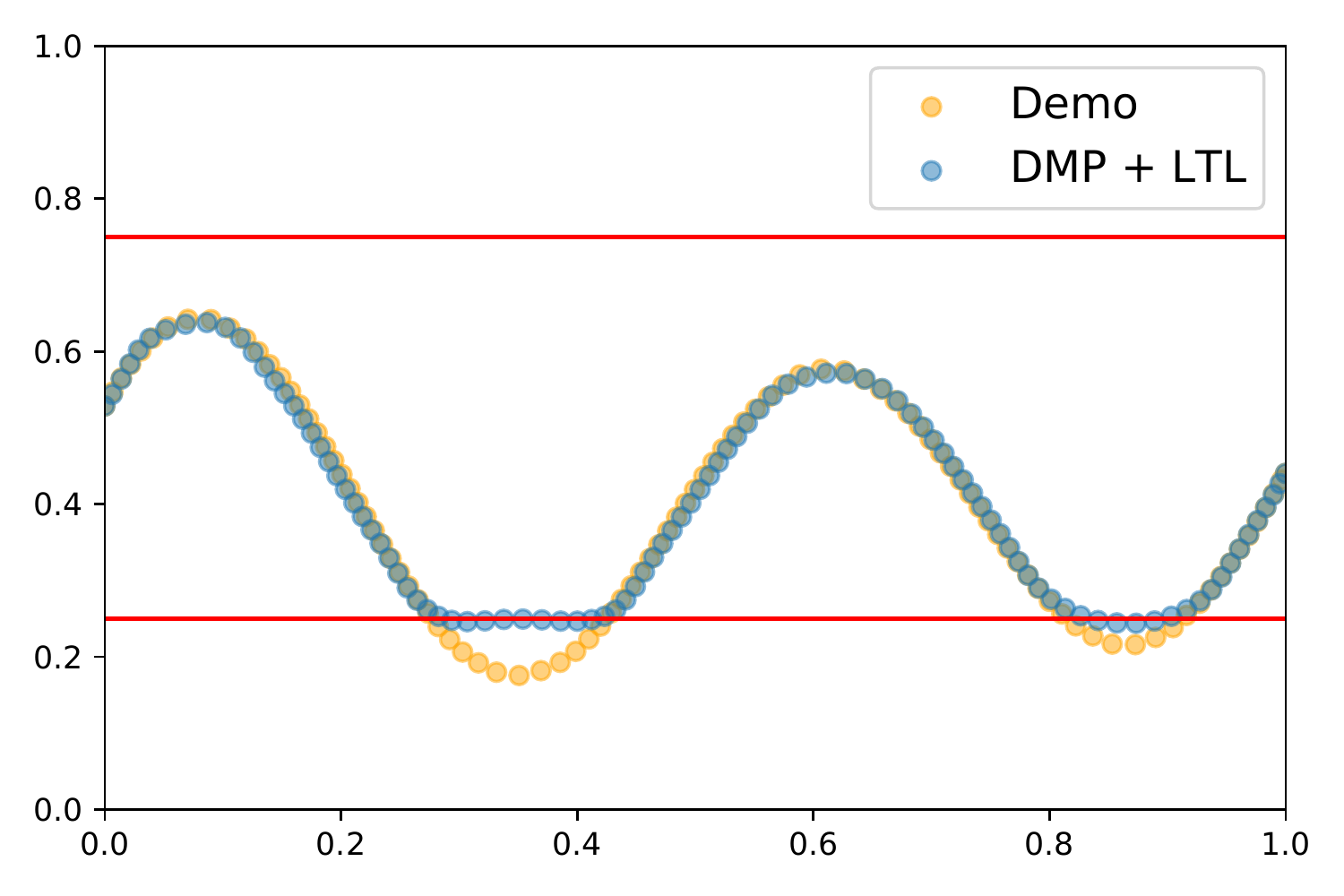}
    \includegraphics[width=0.3\textwidth]{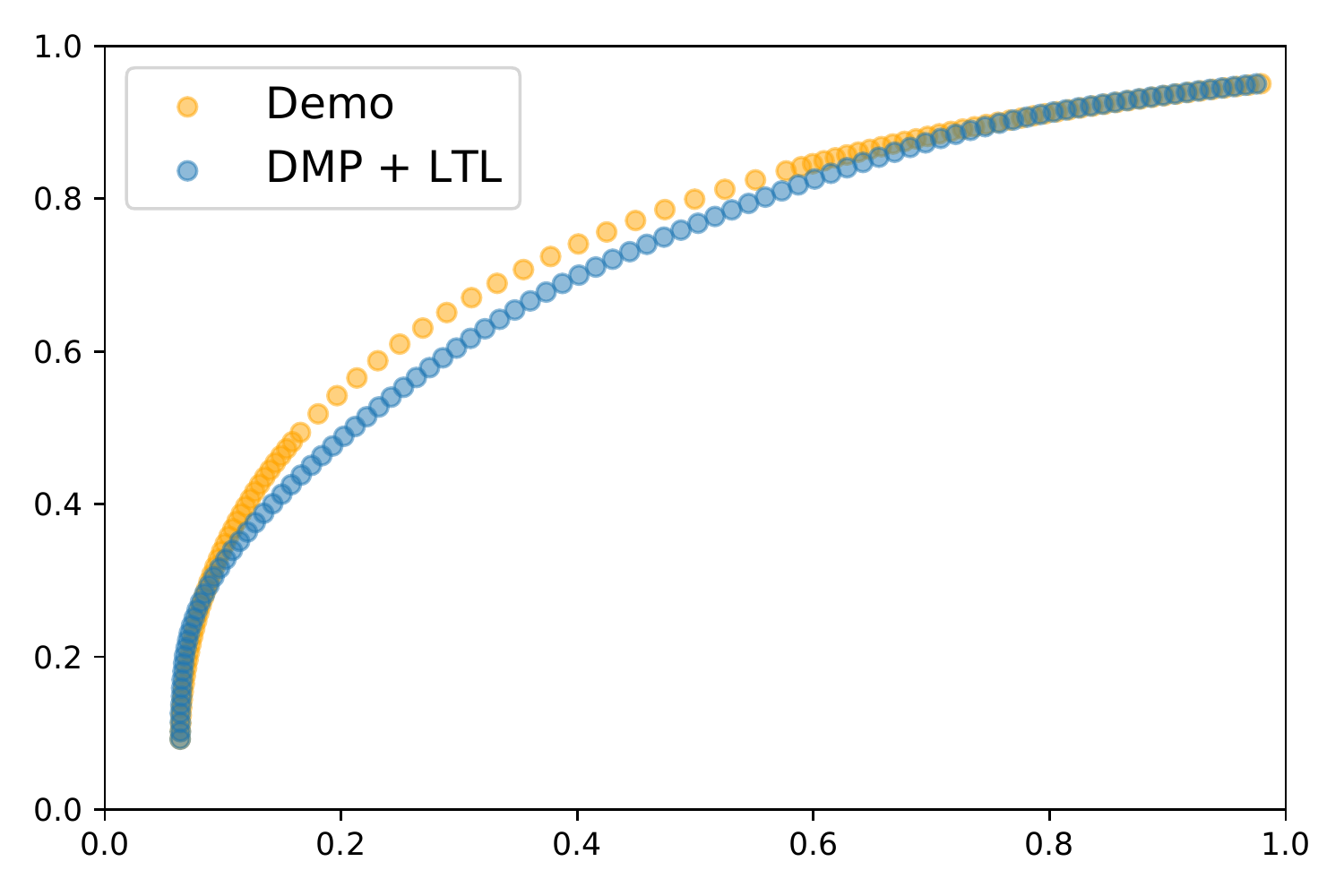}
    
    \caption{The 2-D trajectories for (from left-to-right) avoid, patrol, steady, and slow. The original demonstration is in orange, while the learned trajectory is in blue. Points marked in red provide a visual representation of the associated constraint.}
    \label{fig:single-figure-trajectory}
\end{figure*}

We first consider single-shot demonstrations in a two-dimensional environment. Our goal is to observe whether our model can successfully adapt a base demonstration to one of four \ltl{} constraints, which represent a selection of common robotic movement specifications. These are: Avoiding an obstacle, ensuring we reach a collection of patrol points, keeping movement within fixed bounds, and ensuring we do not move too quickly.

To create our 2-D demonstrations, we generate cubic-spline curves with start/end points in the region $[0, 1]^2$. Our demonstration trajectory $y_i$ then consists of 100 equally-spaced points sampled from the generated curve. We also randomly generate the positions of three ``objects'' within the scene: $o_1$, $o_2$, and $o_3$. These objects may (or may not) be relevant to the underlying task. 

The \ltl{} formulas for the four specifications mentioned above are given below. Here, we write $p = \dmp_{\xin_i, \theta}$ as a shorthand for our learned \dmp{}, $p_y$ to mean the y dimension of a two-dimensional output, and $\dot{p}$ as the current velocity of the \dmp{}: \\

\paragraph{Avoid}
\begin{equation*}
\square \lVert p - o_2 \rVert^2 \geq 0.1
\end{equation*}

\paragraph{Patrol}
\begin{equation*}
(\diamondsuit p = o_2) \wedge (\diamondsuit p = o_3)
\end{equation*}

\paragraph{Steady}
\begin{equation*}
    \square (p_y \geq 0.25 \wedge p_y \leq 0.75)
\end{equation*}

\paragraph{Slow}
\begin{equation*}
    \square \left(\lVert \dot{p} \rVert \leq 0.015 \right)
\end{equation*}

For each task, we randomly generate $100$ curves, varying the start, end and object positions. To train our \dmp{} weights, we use the architecture outlined in figure \ref{fig:architecture-overview}. Our inputs are the start position of the demonstration, the goal position, and the positions of the three generated objects. We skip the object detection step here, as their positions are given to use directly in this section). Each hidden layer contains 256 units. We use the loss term given by equation (\ref{eqn:full-loss-train}) with $\eta=1.0$, $\gamma=0.005$, and train using the Adam optimizer \cite{Kingma2014Adam:Optimization} for 200 epochs with a learning rate of $10^{-3}$. For our \dmp{}, we use the default parameters described by \citet{Hoffmann2009Biologically-inspiredAvoidance}.

\newcommand{\ra}[1]{\renewcommand{\arraystretch}{#1}}
\begin{table}
\centering
\ra{1.3}
\begin{tabular}{@{}lllll@{}}
    \toprule
    & \multicolumn{2}{c}{Unconstrained} & \multicolumn{2}{c}{\textbf{Constrained}} \\
    & $\mathcal{L}_d$ & $\mathcal{L}_c$ & $\mathcal{L}_d$ & $\mathcal{L}_c$ \\   
    \midrule
    Avoid & 0.0046 & 0.0695 & 0.0127 & \textbf{0.0003} \\
    Patrol & 0.0035 & 0.1257 & 0.0228 & \textbf{0.0228} \\
    Steady & 0.0048 & 0.0272 & 0.0079 & \textbf{0.0002} \\
    Slow & 0.0040 & 0.0098 & 0.0045 & \textbf{0.0090} \\
    \bottomrule
\end{tabular}
\caption{The two separate parts of our loss function--- the imitation loss ($\mathcal{L}_d$) and constraint loss ($\mathcal{L}_c$)---on the single-demonstration experiments for the model trained with/without the \ltl{} constraint. Results averaged over 20 iterations.}
\label{table:single-demo-results}
\end{table}

Table \ref{table:single-demo-results} shows the results of our single-demonstration experiments each task. In all cases, the system learns parameters which almost completely minimize the constraint loss $\mathcal{L}_c$, at the cost of slightly increasing in the imitation loss $\mathcal{L}_d$.

Figure \ref{fig:single-figure-trajectory} gives a representative example of the qualitative behavior of our model on each of the task --- The learned trajectory sticks as closely as possible to the original demonstration, except on those areas in which it must diverge to satisfy the additional constraint. Perhaps the most interesting instance of this behaviour (and the one that the model had the most difficulty optimizing) is the ``Slow'' constraint. Here, the learned trajectory appears to ``cut-the-corner'', and turn right early. This pushes adjacent points closer together (minimizing the constraint loss), while keeping as close as possible to each corresponding point in the original demonstration (minimizing the imitation loss).

\begin{figure}
    \centering
    \includegraphics[width=0.6\linewidth]{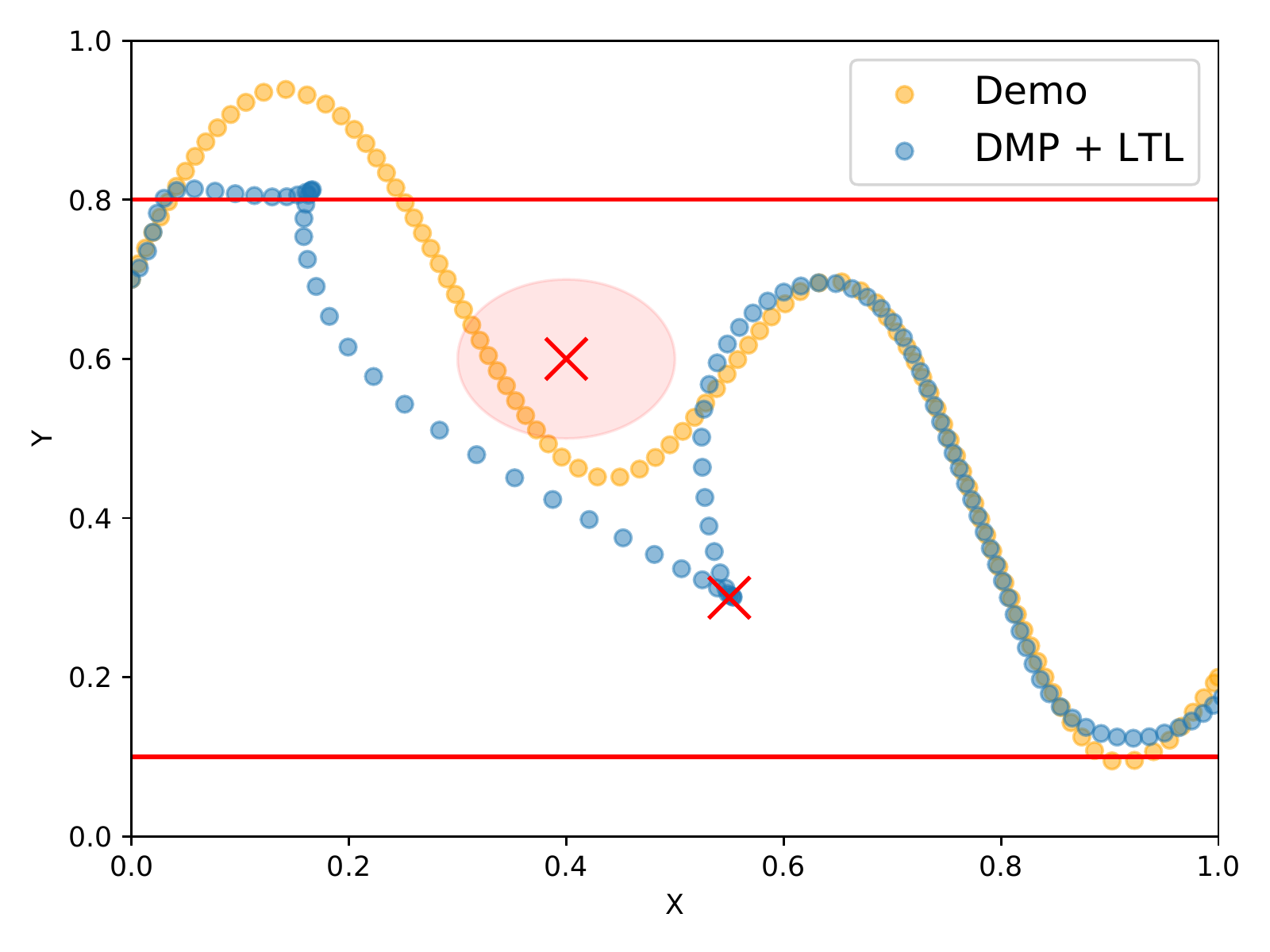}
    \caption{2-D trajectory for a demonstration augmented with avoid, steady, and patrol constraints. Constraints shown in red. This shows the ability of our model to adapt to complex specifications built up from combinations of simpler ones.}
    \label{fig:complex_demo}
\end{figure}

We note here that part of the strength of this approach is that it allows the user to build up increasingly complex specifications declaratively by incrementally composing \ltl{} formulas. For example, figure \ref{fig:complex_demo} shows an example of adding \emph{avoid}, \emph{patrol}, and \emph{steady} specifications incrementally on top of the one base demonstration. The model then computes the best way to minimally augment the original demonstration to satisfy this combination of constraints.

\subsection{Generalized Synthetic Tasks}
\label{sec:generalized-exp}

We next look at whether, given a \emph{batch} of demonstrations, our model can generalize the specified behaviour to unseen inputs. Further, we aim to show that training using the adversarial learning technique outlined in algorithm \ref{alg:full-system} results in more robust satisfaction of the given specification than training using the training data alone.

Our setup is identical to that of section \ref{sec:one-shot-exp}, but instead of optimizing for a single demonstration, we train on a batch of 100 demonstrations per task. We then evaluate the loss on a test set of 20 unseen inputs.

During training, we use an input batch-size of 32, and the inner loop of our adversarial search (lines 6-9) runs for 10 iterations with a robustness domain of $\epsilon = 0.01$. To evaluate the particular contribution of the adversarial steps, we also train a model exclusively on the training data inputs (``Train-Only''). This can be seen as equivalent to setting $\epsilon = 0.0$.

\begin{table}
\centering
\ra{1.3}
\begin{tabular}{@{}llll@{}}
    \toprule
    & Unconstrained & Train-Only & Adversarial \\   
    \midrule
    Avoid & 0.0605 & 0.0272 & \textbf{0.0192} \\
    Patrol & 0.1390 & 0.0413 & \textbf{0.0369} \\
    Steady & 0.0286 & 0.0022 & \textbf{0.0002} \\
    Slow & 0.0089 & 0.0070 & \textbf{0.0069} \\
    \bottomrule
\end{tabular}
\caption{Constraint loss on test set (size 20) for generalized experiments. Models trained without the constraint (Unconstrained), with the constraint but only on the training data (Train-Only), and with adversarial counter-examples (Adversarial).}
\label{table:generalized-results}
\end{table}

Table \ref{table:generalized-results} shows the average constraint loss on the test data for each of our tasks. The model trained without the \ltl{} specifications fails in general to satisfy those constraints on the test set (as the given demonstrations themselves did not satisfy the constraints). When training the system with the constraint loss on the training data alone, the model successfully learns to satisfy the additional specifications to within 2-3 decimal places.

\begin{figure}
    \centering
    \includegraphics[width=\linewidth]{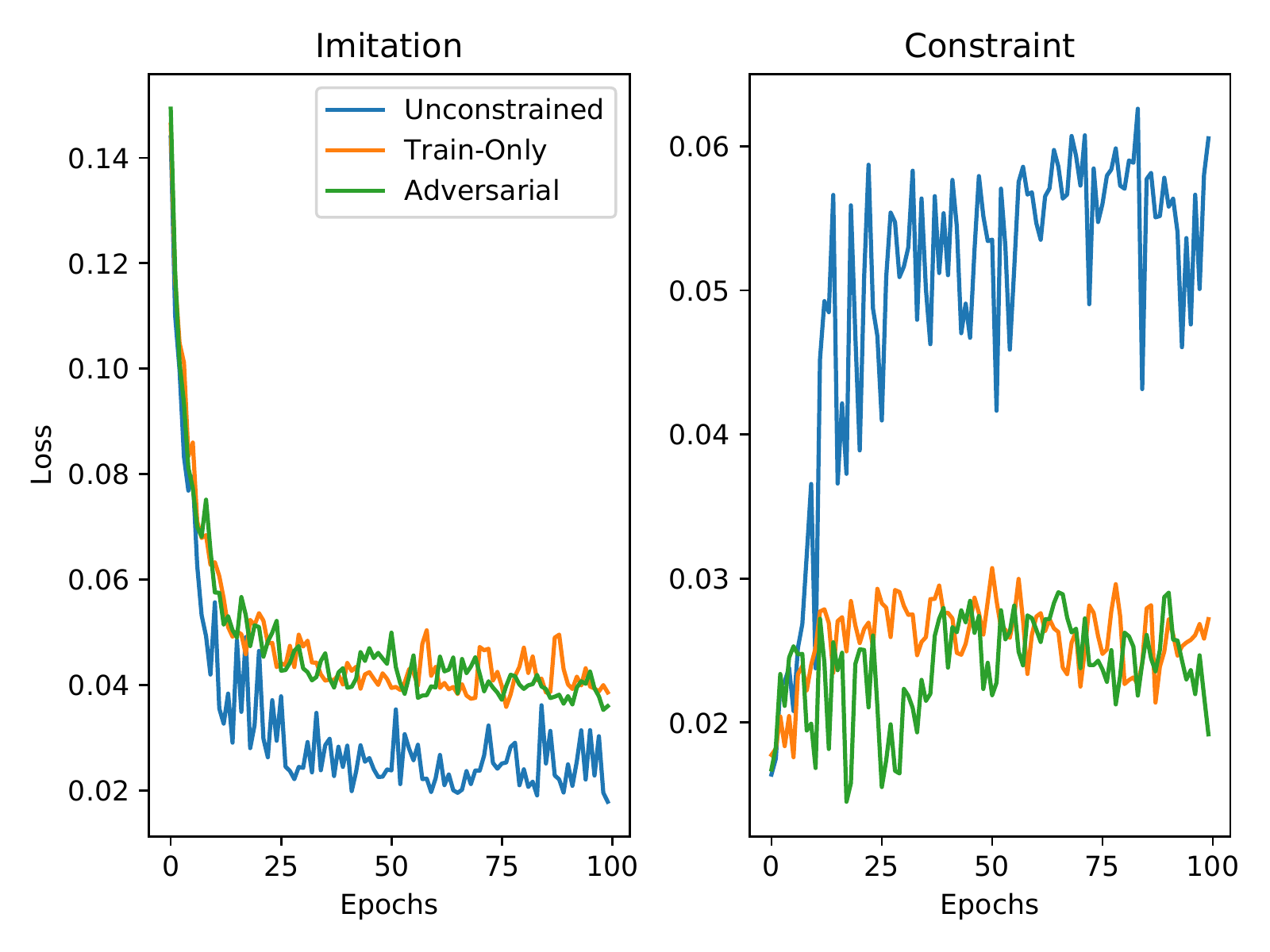}
    \caption{Losses on test set for the ``avoid'' specification.}
    \label{fig:avoid-loss}
\end{figure}

When we train using our adversarial approach, we see minor improvements across each task. Figure \ref{fig:avoid-loss} shows the typical trend during the learning process: In earlier epochs, the adversarial approach is initially able to more aggressively lower the constraint loss, but as the model learns to better imitate the demonstrations, the train-only and adversarial approaches tend to level out. In the later epochs, the train-only and adversarial approaches maintain similar constraint losses, with the adversarial approach typically producing a marginally smaller loss by the final epoch.

An outlier in these results is the \emph{steady} constraint, where the \emph{adversarial} model outperforms the \emph{train-only} approach by an order of magnitude. We hypothesize that this is because the points at which the demonstration trajectories actually violate the boundaries within the training data are sparse, making it difficult for the \emph{train-only} approach to generalize to the unseen test data. In contrast, the adversarial approach proposes new random inputs at every epoch, then pushes these inputs towards places that are maximally likely to violate the boundary conditions. This suggests that the \emph{adversarial} approach might be particularly useful in scenarios where the training data does not fully represent the given specification across the entire space.

\subsection{Experiments on the PR-2 Robot}
\label{sec:pr2-exp}

In this section, we show two applications of our system on real-world manipulation tasks on a PR-2 robot. In the first, we show an example of a pouring task where the user observes the robot imitating their demonstration, then decides to elaborate on their demonstration with an additional \ltl{} specification which captures the underlying constraint they were trying to observe. In the second application, we give an example of a reaching task, where the user later augments the tasks with two completely new specifications that were not present at all in the original demonstration.

We provide demonstrations by teleoperating the PR-2 using a pair of HTC-Vive virtual reality controllers. During teleoperation, we record the six-dimensional end-effector pose of the PR-2's left-hand gripper over time at a rate of 20Hz. This constitutes the demonstrated trajectory. As in previous sections, we use the architecture described by figure \ref{fig:architecture-overview} for learning. The initial scene-object locations are extracted by running the an image from the PR-2's RGB-D camera through off-the-shelf object detection software. To synthesize movement, we then take the end-effector trajectory output by our learned \dmp{} and use an inverse kinematics library\footnote{\url{https://moveit.ros.org/}} to calculate the appropriate joint movements\footnote{The demonstration data used, as well as videos of the final learned system in action, are available at \href{https://sites.google.com/view/ltl-dmp-rss-2020/}{https://sites.google.com/view/ltl-dmp-rss-2020/}}.

\begin{figure}
    \centering
    \includegraphics[width=0.42\linewidth]{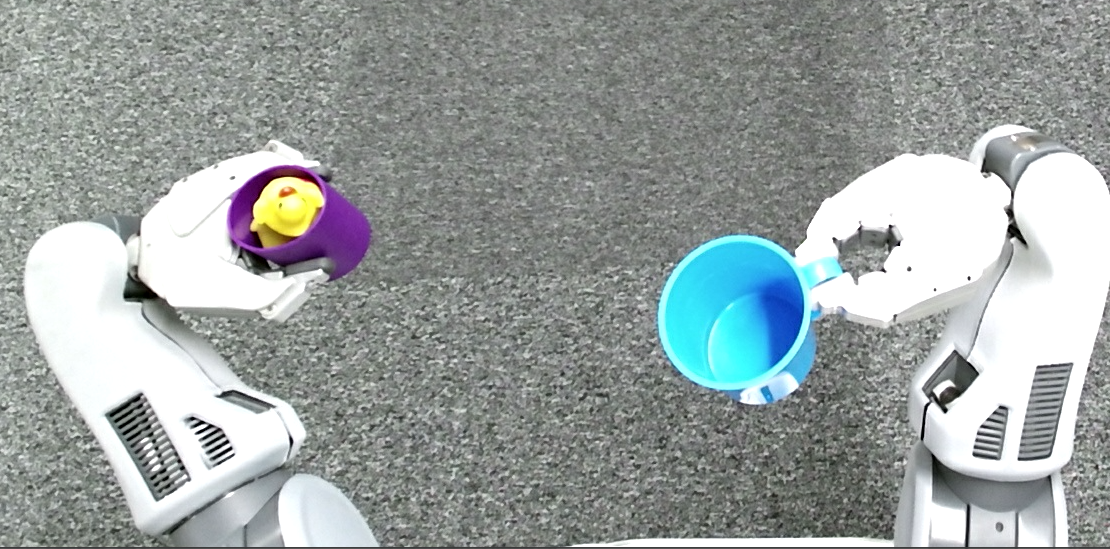}
    \includegraphics[width=0.45\linewidth]{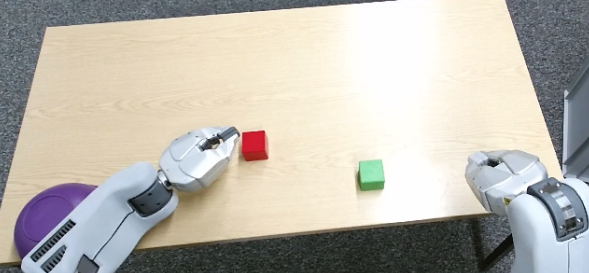}
    \caption{Setup for the pour (left) and reach (right) tasks}
    \label{fig:robot-setup}
\end{figure}

Figure \ref{fig:robot-setup} shows the setup for the two tasks we consider. We describe them in detail below.

\paragraph{Pouring Task}

In this task, the PR-2 is holding a purple cup filled with rubber ducks in one hand, and a blue container in the other. The task is to pour all the ducks from the cup into the container without spilling any. After demonstrating this task several times, we noticed that it was difficult to keep the cup consistently steady via teleoperation, and that we would often tip the cup early before reaching the container (as it is difficult to observe from a distance). We therefore elaborated on our demonstration with the following advice designed to capture an underlying specification: ``Don't tip the cup until you are close to the container''. 

The \ltl{} formula for this specification is given below. As before, we write $p = \dmp{}_{\xin_i, \theta}$ as a shorthand for the pose extracted from our learned \dmp{}, and $p_{y}$ to denote the $y$ dimension of that pose:\\

\begin{align*}
\label{eqn:ltl-tip-early}
\square (\lVert p_{xyz} - o_1 \rVert^2 \geq 0.1\ \wedge\ p_{z} \geq o_{1,z}) \\
\implies (\langle 0,0,-1.0 \rangle \leq p_{rpy} \leq \langle 0.2,0.2,0.0 \rangle)
\end{align*}

\paragraph{Reaching Task}

In this task, we started by demonstrating a simple specification --- Reach from the edge of the table to the red cube. After the demonstration was recorded, we then incrementally add two additional specifications: ``Avoid the purple bowl'', and ``Visit the green cube''. The \ltl{} for this additional specification is given below.

\begin{equation*}
\label{eqn:ltl-bowl-avoid} 
(\square \lVert p_{xyz} - o_2 \rVert^2 \geq 0.2) \wedge (\diamondsuit p_{xyz} = o_ 3)
\end{equation*}

\begin{table}
    \centering
    \ra{1.3}
    \begin{tabular}{@{}lcccc@{}}
        \toprule
        & \multicolumn{2}{c}{Unconstrained} & \multicolumn{2}{c}{Constrained} \\
        & $\mathcal{L}_d$ & $\mathcal{L}_c$ & $\mathcal{L}_d$ & $\mathcal{L}_c$  \\
        \midrule
        Robot-Pour & 0.0033 & 0.0833 & 0.0293 & 0.0038 \\
        Robot-Reach & 0.0059 & 0.1143 & 0.0710 & 0.0104 \\
        \bottomrule
    \end{tabular}
    \caption{Imitation loss ($\mathcal{L}_d$), constraint loss ($\mathcal{L}_c$) for the robotic demonstration tasks.}
    \label{table:robot-experiment-results}
\end{table}

\begin{figure}
    \centering
    \includegraphics[width=\linewidth]{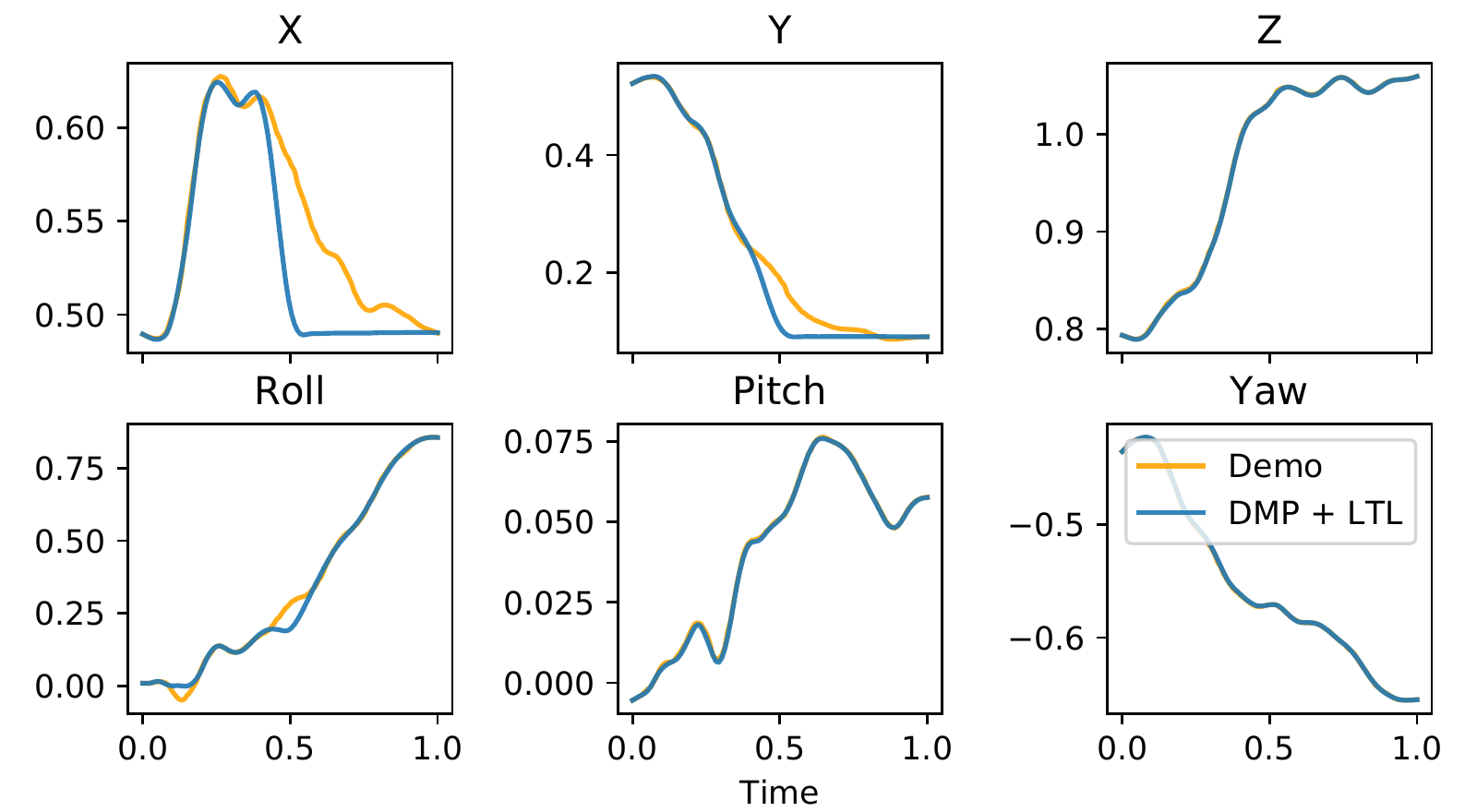}
    \caption{Values of the xyz-rpy dimensions of the robot end-effector over time on the ``cup-pour'' experiment. Rotational axes are normalized between -1 and 1.}
    \label{fig:cup_pour_trajectories}
\end{figure}

As Table \ref{table:robot-experiment-results} shows, our model was able to learn to satisfy the added constraints of both tasks, while keeping close to the initial demonstration. To get a sense of how the model altered the original demonstration, figure \ref{fig:cup_pour_trajectories} shows the value of the six dimensions of the robot's end-effector pose over time on the pouring task. In the roll dimension, we see that the model corrects for an early wobble at around $t=0.1$, and also delays fully tipping the cup at around $t=0.5$. However, it is in the X and Y dimensions that we observe the declarative nature of our specification: Rather than enforce a steady orientation throughout as one might expect, the model instead learns that it can stick closer to the original trajectory by instead negating the left-hand side of the specification implication by bringing the cup closer to the container earlier in the demonstration.

The combined experiments across these three sections show the potential of our method for incrementally learning a task by building additional user specifications on top of an initial physical demonstration.
    
\section{Related Literature}
\label{sec:rel-lit}
The work closest to ours is \citet{Wen2017LearningInformation}, which learns from expert demonstrations with side-information provided by an \ltl{} specification. However, their work frames the problem as inverse-reinforcement learning, where the \ltl{} specification defines a reward function based on the likelihood that the robot will enter a terminating state which satisfies the specification. Other works \cite{Ghosh2018ModelProcesses, li2018policy} also learn a reward function for which an optimal policy will satisfy a given temporal constraint. In contrast, our method uses the \ltl{} specification to directly optimize the weights of a \dmp{}, making it possible to inspect how close the current motion comes to satsifying the given constraint.

Similarly, \citet{Hoffmann2009Biologically-inspiredAvoidance} augment \dmp{}s with a coupling term that avoids obstacles. However, this involves explicitly engineering specialized potential field terms into the underlying differential equations of the \dmp{}. If the specification changes from obstacle avoidance to something else, those equations must be re-engineered. Our method instead allows the user to flexibly alter the behaviour of the \dmp{} by giving an intuitive declarative specification in \ltl{}.

\citet{Kasenberg2018InterpretableSpecifications} consider an inverse problem --- learning an \ltl{} specification \emph{from} a demonstration. As we argued in the introduction however, this approach assumes that the underlying specification for a task is completely learnable from the given demonstrations alone.

Several recent works combine symbolic logic with neural networks. \citet{Xu2018AKnowledge} derive a ``semantic loss'' for logical constraints, which can be used when a network's output layer represents a distribution over a set of boolean variables. Similarly, \citet{Fischer2019DL2:Logic} use their Deep Learning with Differentiable Logic framework to transform propositional logic into differentiable costs for use in classification tasks. They also use an adversarial algorithm to search for constraint-violating counterexamples. We build on such techniques in section \ref{sec:adversarial-learning}, extending their expressiveness to temporal logic for robotic control.

There are many notions of robustness within robotic verification and control. \citet{Tabuada2016RobustLogic} refer to qualitative gradations of failure, such as violating a constraint once versus violating it infinitely. \citet{Fainekos2009RobustnessSignals} convert temporal logic statements into a quantitative metric which measures distance from satisfiability when the robot fails to meet the constraint, and distance from unsatisfiability (robustness) when the robot does meet the constraint. \citet{Donze2010RobustSignals} go further and use metric interval temporal logic to define \emph{time-robustness}. However, the goal of all such work is typically either verification, or high-level planning using a known model of the problem \cite{Raman2012TemporalActions, Kress-Gazit2011CorrectControl, farahani2015robust, Sadigh2015SafeUncertainty}. Our method also encodes a notion of robustness via algorithm \ref{alg:full-system}, but our goals are different. We want to \emph{learn} the low-level control parameters that satisfy the given specification. Consequently, our metric has the added properties of differentiability and having a zero-cost when satisfied. (Arguably, several works which devise metrics of \emph{average robustness} could be also carried over into a network cost function in a similar manner without much effort \cite{lindemann2019robust, mehdipour2019arithmetic}.)

Multiple works combine images, demonstrations and \emph{natural language} \cite{Stepputtis2019ImitationDemonstration, Co-Reyes2019GuidingMeta-learning}. These works encode both images and language as a low-level embedding, then use this embedding to learn a control policy.  Our system does not currently use natural language directly as input. Instead, we use a formal language to specify task constraints. This has the advantage that, unlike the natural language embeddings, the quantitative semantics from equations (\ref{eqn:ltl-diff-semantics-atomic}-\ref{eqn:ltl-diff-semantics-compose}) have an intuitive interpretation, which allows the user to inspect the current model to see how close it is to satisfying the their specification.

\citet{Pervez2017LearningNetworks} and \citet{Pahic2018DeepPrimitives} learn \dmp{}s from images using neural networks. Unlike our paper however, these methods are pure imitation methods --- they assume the given expert demonstration is all that is needed to achieve the underlying specification. Our method has a similar architecture, but elaborates on the initial demonstration using additional specifications encoded in the loss function. 

\citet{Ho2016GenerativeLearning} present an algorithm which alternates between imitating the given expert trajectories, and adversarially learning to discriminate between the current best policy and the expert trajectories. Both our paper and theirs take an adversarial approach to learning, but while their goal is to learn a robust imitation of a given demonstration, ours is ensure that additional \ltl{} specifications are satisfied.

There are analogies between our method and the notion of residual policy learning \cite{Silver2018ResidualLearning}. Here, we start with a fixed control policy for a simpler task, then learn a residual policy to modify the fixed policy for a more complex situation. In this light, we can think of our initial expert demonstration as the fixed policy, and our additional elaboration with the \ltl{} specification as the residual modification.

\section{Conclusions and Future Work} 
\label{sec:conclusion}

In this paper, we presented a quantitative, differentiable loss function which maps to the qualitative semantics of \ltl{}, and an adversarial training regime for enforcing it. Through experiments on a variety of robotic motion tasks, we have shown that our method can successfully elaborate upon a single demonstration with additional specifications, and also ensure that a constraint can be robustly satisfied across unseen inputs when trained on a batch of demonstrations.

For future work, we aim to increase the range of expressible specifications in two ways. The first is to handle time-critical constraints by extending our framework to handle Metric Interval Temporal Logic (\textsc{mitl}) and investigate whether a notion of time-robustness \cite{Donze2010RobustSignals} can be enforced adversarially. The second is to express constraints not just in terms of pre-defined symbols, but also in terms of relational predicates learned directly from data (as in e.g., \cite{Hristov2019DisentangledDemonstration}).

\section*{Acknowledgments}

This work is supported by funding from the Alan Turing Institute, as part of the Safe AI for Surgical Assistance project. Additionally, we thank the members of the \href{http://rad.inf.ed.ac.uk/}{Edinburgh Robust Autonomy and Decisions Group} for their feedback and support.

\bibliographystyle{plainnat}
\bibliography{references}

\end{document}